\documentclass[10pt,twocolumn,letterpaper]{article}

\usepackage{cvpr}      
\usepackage{multirow}
\definecolor{cvprblue}{rgb}{0.21,0.49,0.74}
\usepackage[pagebackref,breaklinks,colorlinks,allcolors=cvprblue]{hyperref}

\def\confName{CVPR}
\def\confYear{2026}

\title{OSGNet with MLLM Reranking @ Ego4D Episodic Memory Challenge 2026}

\author{
Yisen Feng$^{1}$, Leigang Qu$^2$, Haoyu Zhang$^{1\,3}$, Qiaohui Chu$^{1\,3}$, Meng Liu$^{4}$,\\ Xuemeng Song$^{5}$, Weili Guan$^{1}$, Liqiang Nie$^{1}$\\
$^1$Harbin Institute of Technology (Shenzhen) \qquad  $^2$National University of Singapore \\ $^3$Pengcheng Laboratory  \qquad  $^4$Shandong Jianzhu University\\ \qquad  $^5$Southern University of Science and Technology\\
{\tt\small \{yisenfeng.hit, leigangqu, zhang.hy.2019, qiaohuichu8599, mengliu.sdu,} \\ {\tt\small  sxmustc, honeyguan, nieliqiang\}@gmail.com}
}

\begin{document}
\maketitle
\begin{abstract}
In this report, we present our champion solutions for the Natural Language Queries and GoalStep tracks of the Ego4D Episodic Memory Challenge at CVPR 2026. Both tracks require accurately localizing temporal segments from long untrimmed egocentric videos. 
To address these tasks, we propose a reranking-based framework that effectively leverages the strong video-language reasoning capability of multimodal large language model (MLLM) while preserving the efficiency and candidate recall of conventional localization pipelines. 
Specifically, we first obtain a set of candidate segments from existing localization model OSGNet, and then employ MLLM to select the segment that best matches the given query, thereby refining the final prediction.
Ultimately, our method achieved first place in both the Natural Language Queries and GoalStep tracks.
Our code can be found at \url{https://github.com/iLearn-Lab/CVPR25-OSGNet}.
\end{abstract}    
\section{Introduction}
\label{sec:intro}

Egocentric video moment localization~\cite{feng2024objectnlq,feng2025object,feng2025osgnet}, as a fundamental task in egocentric video understanding, underpins a wide range of practical applications~\cite{zhang2021multimodal,10239469}, such as intelligent assistant systems for smart glasses and memory retrieval modules for embodied AI. 
The Ego4D Episodic Memory Challenge~\cite{grauman2022ego4d} includes two video localization tracks. The Natural Language Queries (NLQ) track requires identifying the temporal segment in a long untrimmed egocentric video that answers a given natural language question, while the GoalStep track~\cite{song2024ego4d} focuses on locating the video moment corresponding to a described procedural step. Despite their different query formulations, both tracks require the precise identification of relevant video content based on textual queries.
Despite substantial progress in recent years, conventional approaches~\cite{hou2023groundnlq,liu2018attentive,liu2018cross} built upon pretrained visual and textual features often suffer from limited interpretability and generalization capability, making it difficult to robustly handle diverse queries and complex egocentric scenarios~\cite{pmlr-v235-zhang24aj,zhang2026exo2ego}.

Recent advances in MLLMs~\cite{zhang2026spatial} have opened up new opportunities for egocentric video localization. Recent studies~\cite{wang2026time,li2026universal} have explored using MLLMs to directly generate textual timestamps corresponding to the target moments. However, constrained by the limited context length of MLLMs, long videos can usually be processed only through sparse frame sampling, which may overlook fine-grained temporal evidence and lead to suboptimal localization performance. To address this issue, we propose a reranking-based framework that first employs conventional localization models to retrieve a high-quality set of candidate segments, and then leverages MLLMs to select the segment that best matches the textual query. In this way, MLLMs can focus on distinguishing challenging hard negatives within a compact candidate pool, while the framework also preserves the efficiency advantage of traditional localization methods and substantially reduces the amount of video content that MLLMs need to process.

Building upon this framework, we achieved first place in both the Natural Language Queries and GoalStep tracks of the Ego4D Episodic Memory Challenge at CVPR 2026.
\section{Methodology}
\label{sec:Methodology}

\subsection{Track 1: Natural Language Queries}
\begin{figure*}[t]
  \centering
  \includegraphics[width=\linewidth]{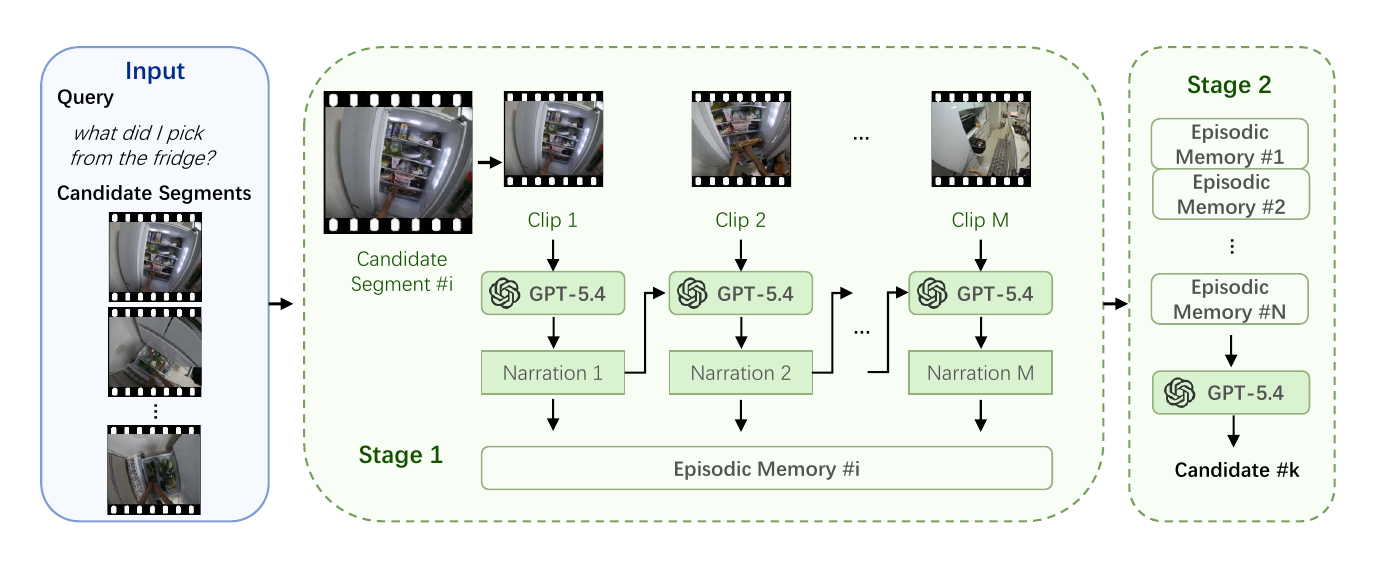}
  \vspace{-4ex}
 \caption{The pipeline of reranking.}
  \vspace{-2ex}
  \label{fig: pipeline}
\end{figure*}
\noindent \textbf{Approach.} We first follow the state-of-the-art method OSGNet~\cite{feng2025osgnet} to generate the top-5 candidate segments for each query, and then use GPT-5.4 to rerank these candidates. Since GPT-5.4 processes visual inputs in the form of images and is subject to a limit on the number of input images, we decompose the reranking procedure into two stages. As shown in figure~\ref{fig: pipeline}, in the first stage, each candidate segment is divided into 20-second clips, from which frame sequences are sampled at 1 FPS and sequentially fed into GPT-5.4 to produce textual narrations. These narrations are accumulated to construct an episodic memory for each candidate segment. In the second stage, we collect the episodic memories of all candidates and prompt GPT-5.4 to perform reasoning over them, ultimately selecting the segment that best matches the given natural language query.

\begin{table}[t]
\caption{Performance on NLQ test split.}

  \vspace{-2ex}
  \label{tab: NLQ}
  \centering
  \begin{tabular}{lcccc}
    \toprule
    \multirow{2}{*}{Method} & \multicolumn{2}{c}{R@1}&\multicolumn{2}{c}{R@5}\\
       & 0.3 & 0.5 & 0.3 & 0.5\\
    \midrule

     OSGNet~\cite{feng2025object} &21.63&\textbf{15.52}&41.74&31.85\\
     OSGNet w/ rerank &\textbf{21.78} &15.44 &\textbf{41.74} &\textbf{31.85}  \\
    \bottomrule
  \end{tabular}
\end{table}

\noindent \textbf{Results.} As shown in Table~\ref{tab: NLQ}, reranking partial samples (630 items) slightly improves OSGNet from 21.63\% to 21.78\% in R@1, mIoU=0.3, while the performance at R@1, mIoU=0.5 changes from 15.52\% to 15.44\%. The R@5 results remain unchanged at both IoU thresholds. Overall, the improvement on NLQ is relatively limited, which may be affected by the false negative segments, making the benefit of reranking less fully reflected by the metrics.

\noindent\textbf{Case Analysis.}
\begin{figure}[t]
  \centering
  \includegraphics[width=\linewidth]{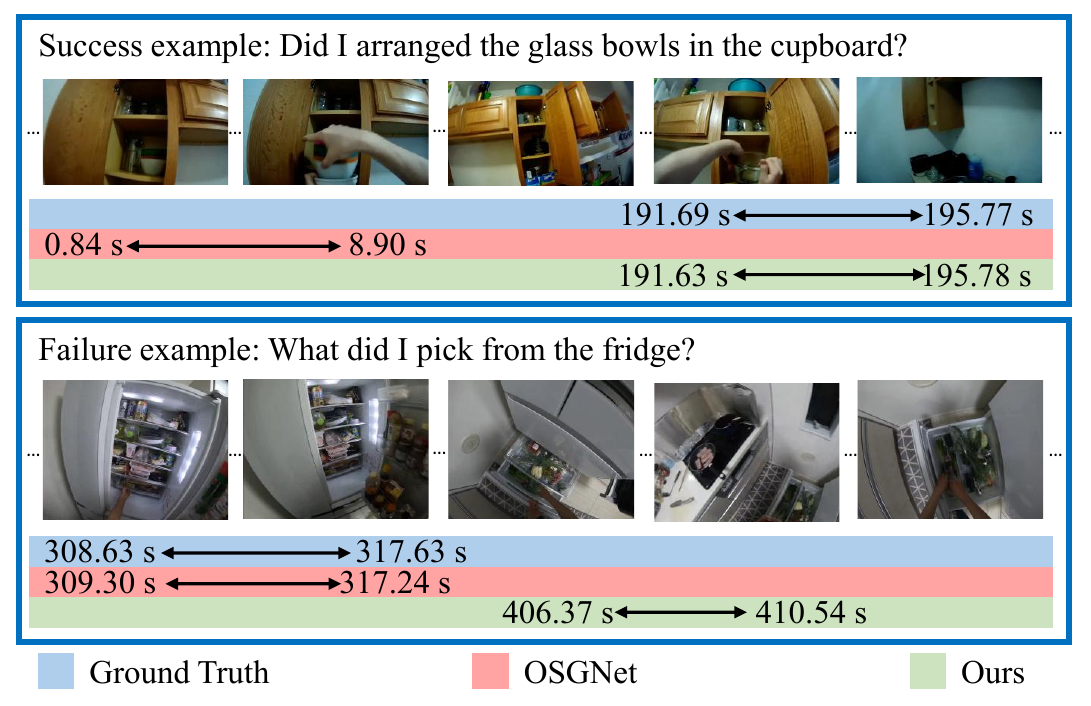}
  \vspace{-4ex}
  \caption{Two examples on the validation set of NLQ.}
  \vspace{-2ex}
  \label{fig: NLQ}
\end{figure}
As shown in Figure~\ref{fig: NLQ}, in the successful example, OSGNet initially ranks a segment showing the arrangement of ceramic bowls as the top prediction, which is visually similar but semantically inconsistent with the queried object. By comparing the candidate segments with stronger visual-language reasoning, GPT-5.4 successfully reranks the correct segment to the first position. This example demonstrates the advantage of MLLM-based reranking in distinguishing hard negatives with subtle differences.
In the failure example, the reranked prediction does not match the annotated ground-truth segment, but it still shows a plausible moment that satisfies the query. This helps explain the limited improvement of reranking under the current evaluation protocol.
\subsection{Track 2: GoalStep - Step Grounding}
\begin{table}[t]
\caption{Performance on GoalStep test split.}

  \vspace{-2ex}
  \label{tab: GoalStep}
  \centering
  \begin{tabular}{lcccc}
    \toprule
    \multirow{2}{*}{Method} & \multicolumn{2}{c}{R@1}&\multicolumn{2}{c}{R@5}\\
       & 0.3 & 0.5 & 0.3 & 0.5\\
    \midrule

     OSGNet~\cite{feng2025object} &55.31&47.82&80.12&74.93\\
     OSGNet w/ rerank &\textbf{55.39} &\textbf{47.91} &\textbf{80.12} &\textbf{74.93}  \\
    \bottomrule
  \end{tabular}
\end{table}
\begin{table*}[t]
\centering
\caption{Official leaderboard results of the GoalStep track.}
\label{tab:goalstep_leaderboard}
\begin{tabular}{c|l|cccc|c}
\toprule
\# & Participant 
& R@1@0.3 & R@1@0.5 
& R@5@0.3 & R@5@0.5 
& Mean R@1 \\
\midrule
1 & yisen\_feng  & \textbf{63.02} & \textbf{54.21} & \textbf{80.12} & \textbf{74.93} & \textbf{58.61} \\
2 & andreazenotto & 56.27 & 40.20 & 77.39 & 61.38 & 48.24 \\
3 & kaname06      & 53.39 & 45.43 & 79.91 & 73.59 & 49.41 \\
4 & yuki11        & 52.41 & 44.55 & 79.09 & 72.64 & 48.48 \\
5 & willy06	      &	36.47 &	22.65 &	64.28 &	45.61 &	29.56 \\
\bottomrule
\end{tabular}
\end{table*}

\noindent \textbf{Approach.} We adopt the same reranking pipeline as in the NLQ track, using OSGNet to generate top-5 candidate segments and GPT-5.4 to select the segment that best matches the given description.

We further exploit the sequential prior of the GoalStep dataset during post-processing. For a video containing an ordered sequence of \(K\) step queries \(\{q_i\}_{i=1}^{K}\), we select one candidate segment for each query. Let \(\mathrm{start}_i\) denote the start timestamp of the selected segment for the \(i\)-th query, and let \(r_i\) denote the rank of this selected candidate in the candidate list, where a smaller rank indicates a higher preference from the localization model.

Based on the sequential prior of the GoalStep dataset, we encourage the selected segments to satisfy a monotonically increasing start-time constraint. For each adjacent query pair $(q_i, q_{i+1})$, we define a start-time violation penalty as
\begin{equation}
\mathrm{penalty}^{\mathrm{start}}_i
=
\max(0,\, \mathrm{start}_i - \mathrm{start}_{i+1}).
\end{equation}

We then search for the set of candidate selections that minimizes the overall cost.
\begin{equation}
\mathrm{cost}
=
\sum_{i=1}^{K} r_i
+
\sum_{i=1}^{K-1}
\mathrm{penalty}^{\mathrm{start}}_i.
\end{equation}

\noindent \textbf{Results.} 
As shown in Table~\ref{tab: GoalStep}, the reranking strategy brings consistent improvements on GoalStep. After applying reranking to partial samples (850 items), the performance increases from 55.31\% to 55.39\% in R@1, mIoU=0.3, and from 47.82\% to 47.91\% in R@1, mIoU=0.5. The R@5 results remain unchanged at both IoU thresholds. These results indicate that MLLM-based reranking can further refine the predictions of OSGNet and improve the selection of the most relevant candidate segment. 

Beyond reranking, our post-processing substantially improves the final performance and enables our method to achieve first place in the GoalStep track, as shown in Table~\ref{tab:goalstep_leaderboard}. It is worth mentioning that the annotations or evaluation metrics of the challenge may have changed in this year, the results of NLQ and GoalStep are not directly comparable with those reported in previous years.

\noindent\textbf{Case Analysis.}
\begin{figure}[t]
  \centering
  \includegraphics[width=\linewidth]{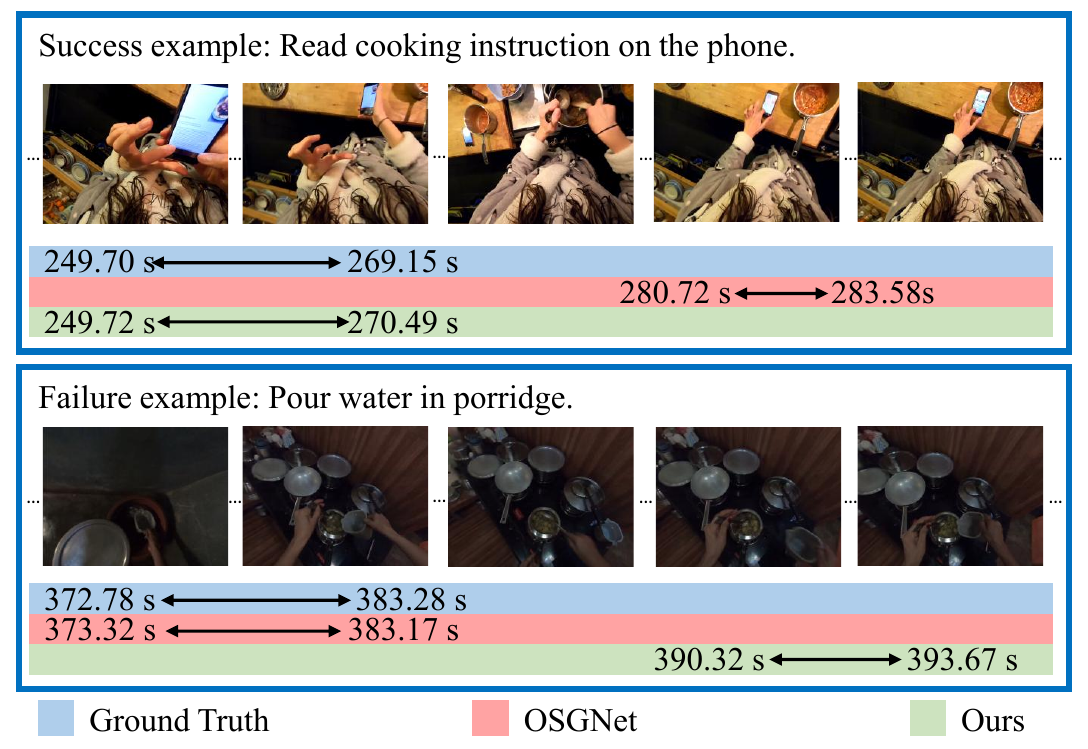}
  \vspace{-4ex}
  \caption{Two examples on the validation set of GoalStep.}
  \vspace{-2ex}
  \label{fig: GS}
\end{figure}
As shown in Figure~\ref{fig: GS}, the successful case illustrates the benefit of reranking for resolving fine-grained semantic ambiguity. OSGNet initially selects a segment where the person briefly checks a phone notification, which involves the same object but does not match the intended procedural step. GPT-5.4 instead promotes the candidate showing the person reading cooking instructions on the phone to the top rank, leading to a correct prediction.
In the failure case, although the reranked prediction does not coincide with the annotated ground-truth interval, both segments depict the action of pouring water into porridge. This again highlights the influence of latent positives on the reported localization performance.

\section{Conclusion}

This report presents our solutions for the NLQ and GoalStep tracks of the Ego4D Episodic Memory Challenge at CVPR 2026. We explore a reranking-based localization framework that combines the efficiency of conventional grounding models with the strong video-language reasoning ability of MLLMs. Ultimately, our approach achieved first place in both tracks.
{
    \small
    \bibliographystyle{ieeenat_fullname}
    \bibliography{main}
}


\end{document}